\documentstyle[aclap]{article}

\newcommand{\lline}{\noindent\rule{\textwidth}{0.25mm}}

\newenvironment{bullist}
    {\begin{list}{$\bullet$}
	{\parsep 0pt \itemsep 0pt \setlength{\rightmargin}{\leftmargin}}}%
    {\end{list}}

\title{Two Sources of Control over the Generation of Software
Instructions\thanks{\hspace*{1mm} This work is partially supported by the Engineering and
Physical Sciences Research Council ({\sc epsrc}) Grant \mbox{J19221},  by
{\sc bc/daad arc} Project 293,  by the Commission
of the European Union Grant \mbox{{\sc lre}-62009}, and by the Office
of Naval Research Grant \mbox{N00014-96-1-0465}.}}
\author{Anthony Hartley\\
        Language Centre\\
        University of Brighton, Falmer\\
        Brighton BN1 9PH, UK\\
        {\tt afh@itri.bton.ac.uk}\And
        C\'ecile Paris\thanks{\hspace*{1mm} Starting this fall,
Dr. Paris' address will
be CSIRO, Division of Information Technology, Sydney Laboratory, Building
E6B, Macquarie University Campus, North Ryde, Sydney, NSW 2113,
Australia.}\\
	Information Technology Research Institute\\
        University of Brighton,	Lewes Road\\
	Brighton  BN2 4AT, UK\\
	{\tt clp@itri.brighton.ac.uk}}

\begin{document}
\maketitle
\bibliographystyle{acl}
\section{Introduction}

Our work addresses the generation of software manuals in French and
English, starting from a semantic model of the task to be documented
\cite{drafter-ijcai95}.
Our prime concern is to be able to exercise control over the mapping
from the task model to the generated text.  We set out to establish
whether the task model alone is sufficient to control the linguistic
output of a text generation system, or whether additional control is
required. In this event, an obvious source to explore is the
communicative purpose of the author, which is not necessarily constant
throughout a manual.  Indeed, in a typical software manual, it is
possible to distinguish at least three sections, each with a different
purpose: a tutorial containing exercises for new users, a series of
step-by-step instructions for the major tasks to be accomplished, and
a ready-reference summary of the commands.

We need, therefore, to characterise the linguistic expressions of the
different elements of the task model, and to establish whether these
expressions are sensitive or not to their context, that is, the
functional section in which they appear.  This paper presents the
results of an analysis we conducted to this end on a corpus of
software instructions in French.

\section{Methodology}

The methodology  we employed is similar to that endorsed by
\cite{Biber95}.  It is  summarised as follows:
\begin{enumerate}
\item Collect the texts and note their situational characteristics. We consider two
such characteristics:   task structure and communicative purpose.
\item Identify the range of linguistic features to
be included in the analysis;
\item Code the corpus in terms of the selected features;
\item Compute the frequency count of each linguistic feature;
\item Identify co-occurrences between linguistic features and the
situational characteristics under consideration. 
\end{enumerate}

We first carried out a classical sublanguage analysis on our corpus as
a whole, without differentiating between any of the situational
characteristics \cite{drafter-wp14}.  This initial description was
necessary to give us a clear statement of the linguistic potential
required of our text generator, to which we could relate any
restrictions on language imposed by situational variables.  Thus we
can account for language restrictions by appealing to general
discourse principles, in keeping with the recommendations of
\cite{Kittredge95} and \cite{Biber95} for the definition of
sublanguages.

We then  correlated  task  elements with grammatical
features.    Finally,  where linguistic realisation was
under-determined by task structure alone, we  established whether the 
communicative purpose provided more discriminating control
over the  linguistic resources available.

\section{Linguistic Framework: Systemic Functional Linguistics}
Our analysis   was      carried   out  within     the  framework  of 
Systemic-Functional Linguistics ({\sc sfl})
\cite{halliday78:systemics,halliday85a:systemics} which views
language as  a  resource for   the creation  of meaning.    {\sc  sfl}
stratifies meaning into  context and language.    The
strata   of  the linguistic 
resources are  organised   into   networks of  choices,  each   choice
resulting in a different meaning {\em realised\/} (i.e., expressed) by
appropriate structures.     The    emphasis  is     on   {\em
paradigmatic\/} choices, as opposed to {\em syntagmatic\/} structures.
Choices made  in each stratum  constrain the choices available  in the
stratum beneath.  Context thus constrains language.

This framework was chosen for several reasons.  First, the
organisation    of linguistic  resources  according to  this
principle  is well-suited to  natural  language generation, where  the
starting point is necessarily a communicative goal, and the task is to
find the most appropriate expression for the intended meaning
\cite{MatthiessenBateman91}. 
Second, a functional perspective offers an advantage for
multilingual text generation, because of its ability to achieve a
level of linguistic description which holds across languages more
effectively than do structurally-based accounts.
The approach has
been shown capable of  supporting the sharing of linguistic  resources
between languages as structurally distinct as English and Japanese
\cite{Bateman-etal90-prototyping,Bateman-etal91-multi-penang}.  It
is therefore reasonable to expect that at least the same degree of
commonality of description is achievable between English and French
within this framework.
Finally, {\sc kpml} \cite{kpml-manual}, the tactical generator we
employ,  is based on {\sc sfl}, and 
it is thus appropriate for us to characterise the corpus in terms
immediately applicable to our generator.

\section{Coding features}
Our lexico-grammatical coding was
done using the networks  and   features of the Nigel grammar 
\cite{halliday85a:systemics}.  
We focused on four main concerns, guided  by  previous work  on
instructional texts, e.g.,
\cite{Lehrberger86,Plum90-eda-text-analysis,Ghadessy93,kosseim-lapalme94}.
\begin{bullist}
\item {\em Relations between processes\/}: to determine whether textual
cohesion was achieved through conjunctives or through relations
implicit in the task structure elements.   Among the features
considered were  clause dependency and conjunction type.

\item  {\em Agency\/}: to see whether the actor performing or enabling
a particular action is clearly identified, and whether the reader is
explicitly addressed.  We coded here for  features such as voice and
agent types.

\item {\em Mood, modality and polarity\/}: to  find out the extent
to which  actions  are  presented  to the reader  as  being desirable,
possible,  mandatory,  or prohibited.   We   coded for  both  true and
implicit negatives, and  for both personal and impersonal  expressions
of modality.

\item {\em Process types\/}: to see how the domain is construed in
terms of actions on the part of the user and the software. We coded
for sub-categories of material, mental, verbal and relational processes.
\end{bullist}

\section{The Corpus}
The  analysis was conducted on  the French version of the
Macintosh MacWrite manual
\cite{MacWrite-french}. The manual is derived  from an English  source
by a process of {\em 
adaptive translation\/}  \cite{Sager93}, i.e., one which localises the
text to the expectations of the target readership.
The fact that the translation is adaptive
rather than literal gives us confidence in using this manual for our
analysis.\footnote{We would  have preferred to  use a manual which
originated in French to exclude all possibility of interference from a
source language, but this proved impossible.  Surprisingly, it appears
that large French companies often have
their documents  authored in English by francophones and 
subsequently translated into  French.  One large French software house
that we  contacted does author   its documentation in French, but  had
registered considerable customer dissatisfaction with its quality.  We
decided, therefore, that their  material would  be unsuitable for  our
purposes.}  Furthermore, we know that Macintosh documentation undergoes
thorough local quality control.  It certainly conforms to the
principles of good documentation established by current research on
technical documentation and on the needs of end-users, e.g.,
\cite{Carroll-94:tech-writing,Hammond94:tech-writing}, 
in that it supplies clear and concise information for the task at hand.
Finally, we have been assured by French users of the software that
they consider this particular manual to be well written and to bear no
unnatural trace of its origins.  

\begin{figure*}[t]
\lline \\
\begin{tabbing}
nnnnnnnnnnnn\= task nn\= nnnnnn \kill
{\bf Goals:}\> La s\a'{e}lection\\
\> {\em Gloss:} \> Selection \\[0.1in]

\> Pour s\a'{e}lectionner un mot, (faites un double-clic sur le mot)\\
\> {\em Gloss:} \> To select a word, (do a double-click on the
word)\\[0.2in]

{\bf Functions:}\> (Fermer --) Cet article permet de fermer une fen\^etre activ\a'{e}e \\
\> {\em Gloss:} \> (Close --) This command enables you to close the
active window \\[0.2in]

{\bf Constraints:}\> Si vous donnez \a`{a} votre document le titre d'un
document d\a'{e}j\a`{a} existant, (une zone de \\
\> dialogue appara\^{\i}t)\\
\> {\em Gloss:} \> If you give your document the title of an
existing document, (a dialog box \\
\>\>appears)\\[0.2in]

{\bf Results:}\> (Choisissez Coller dans le menu Edition -- )  Une
copie du contenu du presse-papiers appara\^{\i}t\\
\> {\em Gloss:} \> (Choose Paste from the Edit menu --) A copy of
the content of the clipboard appears\\[0.2in]

{\bf Substeps:}\> Fermez la fen\^etre Rechercher\\
\> {\em Gloss:} \> Close the Find window\\[0.1in]
\> Ensuite, on ouvre le document de destination\\
\> {\em Gloss:} \> Next, one opens the target document\\
\end{tabbing}
\caption{Examples of task element expressions}
\lline
\label{ex-task}
\end{figure*}
Technical manuals   within   a specific  domain    constitute   a
sublanguage, e.g.,
\cite{Kittredge82,Sager-Dungworth-McDonald}.  An important
defining property of a sublanguage is that of closure, both lexical
and syntactic. Lexical closure has been demonstrated by, for example,
\cite{Kittredge87-nirenburg}, who shows that after as few as 	the first 2000
words of a sublanguage text, the number of new word types increases
little if at all.   Other work, e.g., 
\cite{Biber88,Biber89} and
\cite{Grishman86} illustrates the property of syntactic closure,
which  means that generally available  constructions just do not occur
in this  or  that  sublanguage. In the   light  of these results,  we
considered   a   corpus of 15000  words to be 
adequate for our purposes, at least for an initial analysis.

The MacWrite manual is organised into three chapters, corresponding to
the three different sections identified earlier: a tutorial, a
series of step-by-step 
instructions for the major word-processing tasks, and a
ready-reference summary of the commands.  We omitted 
the tutorial because the generation of such text is not our
concern,  retaining the other two chapters which provide the user
with generic instructions for performing relevant tasks, and
descriptions of the commands available within MacWrite.  The overlap
in information between the two chapters offers 
opportunities to observe   differences in the
linguistic expressions of the same task structure elements in different contexts.

\section{Task Structure}

Task structure is constituted by five types of task elements, which we
define below.  We used the notion of task structure element both as a
contextual feature for the analysis and to determine the segmentation
of the text into units.  Each unit is taken to be the expression of a
single task element.

Our  definition of the task  elements   is based on  the
concepts and relations  commonly chosen to represent  a task structure (a  goal
and    its   associated   plan), e.g., 
\cite{fikes71:planning,sacerdoti77:planning},  and  on related research, e.g.,
\cite{kosseim-lapalme94}.   Our generator produces instructions from
an underlying semantic knowledge base which uses this representation
\cite{drafter-ijcai95}. 
To generate an instruction for performing a task is to chose some task
elements to be expressed and linearise them so that they form a
coherent set for a given goal the user might have.  We distinguish the
following elements, and provide examples of them in
Figure~\ref{ex-task}:\footnote{The text in parentheses in the Figure
is part of the linguistic context of the task element rather than the
element itself.}

\begin{description}
\item [goals:] actions that  users will adopt as goals
and which motivate the use of a plan. 

\item [functions:]  actions that represent the functionality of an
interface object (such as  a  menu item).  A  function is
closely related to a goal, in that it is also  an action that the user
may want to  perform.  However, the function  is accessed  through the
interface object, and not through  a plan.

\item [constraints and preconditions:] states which must hold before a
plan can be  employed  successfully.  The domain  model  distinguishes
constraints (states which  cannot  be achieved through  planning)  and
preconditions (states which can be  achieved through planning).  We do
not make this distinction in the linguistic analysis and regroup these
related   task structure  elements  under one  label.    We decided to
proceed in this way  to determine at  first how constraints in general
are expressed. Moreover, it is not  always clear {\em from the text\/}
which type of constraint  is expressed.  Drawing too fine distinctions
in the corpus  analysis at this point,  in  the absence of a  test for
assigning a unit to one of these constraint types, would have rendered
the results of the analysis more subjective and thus less reliable.

\item [results:] states which arise as planned or  unplanned effects 
of carrying  out a  plan.  While   it might  be important  to separate
planned   and unplanned effects in   the underlying representation, we
again abstract over them in the lexico-grammatical coding.

\item [sub-steps:]  actions  which contribute to 
the execution of the plan. If the sub-steps  are not primitive, they
can themselves be achieved through  other plans.
\end{description}

\section{The Coding Procedure}

No tools exist to automate a functional  analysis of text, which makes
coding a large body of text a time-consuming task.  We first performed
a  detailed  coding of units  of texts  on   approximately 25\% of the
corpus, or about  400 units,\footnote{The authors  followed guidelines
for identifying   task   element units which  had  yielded  consistent
results when  used by students coding  other corpora.} using the {\sc
wag} coder \cite{Mick-Wag}, a  tool designed to facilitate a
functional analysis. 

We then used a public-domain concordance  program, MonoConc \cite{monoconc}, to verify
the representativeness of the results.  We enumerated the realisations
of  those features that the  first analysis  had  shown as marked, and
produced KWIC\footnote{Key Word In Context} listings  for each set  of
realisations.   We found that 
the second analysis corroborated the results  of the first, consistent
with the nature of sublanguages.

\section{Distribution of Grammatical Features over Task Structure and
Communicative Purpose}
\begin{figure*}[t]
\lline
\begin{tabbing}
nnnnnnnnn\=modal-systemnnnnn\= functionnnnnn\= resultnnnnn\= constraintnnnn\=
goalnnnnn \= substep \kill 

\> \> {\bf Function} \> {\bf Result} \> {\bf Constraint} \> {\bf Goal} \> {\bf
Substep} \\ 
\> {\bf Modal-System} \\[0.12in]
\> modal        \>  0\%  \>  1\% \>   0\% \> 24\% \> 16\%       \\ 
\> non-modal    \> 100\% \> 99\% \> 100\% \> 76\% \> 84\%       \\ [0.1in]
\> {\bf polarity} \\
\> positive        \>  100\%  \>  90\% \>   68\% \>   97\% \> 97\%       \\ 
\> negative        \>  0\% \>     10\% \>   32\% \>    3\% \>  3\% \\[0.1in]
\> {\bf mood-system} \\
\> declarative        \>  100\%  \>  100\% \>  100\% \>   100\% \> 24\%       \\
\> imperative         \>  0\% \>       0\% \>    0\% \>     0\% \> 76\%       
\end{tabbing}
\caption{Selective realisations of task  elements}
\lline
\label{task-element-realisation}
\begin{tabbing}
nnnnnnnnnn\=nnnnnnnnnnnnnnn\= reader-referencennnn\= procedurennnn \= elaboration \kill
\>     \> {\bf Ready-Reference} \> {\bf Procedure} \> {\bf Elaboration} \\[0.12in]
\> Sub-step   \> 37\%                \> 77\%            \> 42\%       \\
\> Goal       \> 11\%                \> 23\%            \> 14\%       \\
\> Constraint \> 10\%                \>  0\%            \> 14\%       \\
\> Result     \> 23\%                \>  0\%            \> 27\%       \\
\> Function   \> 11\%                \>  0\%            \>  3\%       
\end{tabbing}
\caption{Distribution of task structure elements over genres}
\lline
\label{cross-ref}
\end{figure*}
We examined the correlations between lexico-grammatical realisations
and  task  elements and communicative purpose. The results are best
expressed using  tables generated by {\sc wag}: given any system, {\sc wag} splits
the codings into a number of sets, one for each feature in that
system. Percentages and means are computed, and the sets are compared
statistically, using the standard T-test.  {\sc wag} displays the
results with an indicator of how statistically significant a value is
compared to the combined means in the other sets.  The counts were all
done using the local mean, that is, the feature count is divided by
the total number of codings which select that feature's system.  Full
definitions of the features can be found in
\cite{halliday85a:systemics,upper-model:penman}.

In some cases, the type of task element is on its own sufficient to
determine, or at least strongly constrain, its linguistic
realisation. 
The limited space available here allows us to  provide only 
a small number of examples, shown in
Figure~\ref{task-element-realisation}. We see that the use of modals
is excluded in the expression of 
function, result and constraint, whereas goal and sub-step do admit
modals.  As far as the polarity system is concerned, negation is
effectively ruled out for function, goal and substep. Finally, with
respect to the mood system, only substep can be realised through
imperatives. 

In other  cases, however, we observe a  diversity of realisations.  We
highlight here three cases:  modality in goal, polarity in constraint,
and mood in substep.  In such cases, we must  appeal to another source
of  control over the apparently  available  choices. We have looked to
the  construct  of  {\em   genre\/}  \cite{Martin92} to provide   this
additional control, on two grounds: (1) since genres are distinguished
by their communicative purposes, we  can view  each of the  functional
sections already    identified  as a  distinct   genre;  (2) genre  is
presented as controlling text structure  and realisation.  In Martin's
view, genre   is defined  as  a staged,  goal-oriented  social process
realised through register, the context of situation,  which in turn is
realised  in language  to   achieve the goals   of a  text.  Genre  is
responsible for  the selection  of a  text structure in  terms of task
elements.    As part of the     realisation process, generic   choices
preselect a  register  associated  with particular  elements  of  text
structure, which in turn preselect lexico-grammatical features.
The coding of our  text in terms genre and task elements thus allows
us to establish the role played by genre in the realisations of the
task elements. It will also allow us   to determine  the  text
structures appropriate in each 
genre, a study we are currently  undertaking.  This is consistent with
other accounts of  text  structure  for text generation  in  technical
domains, e.g., \cite{mckeown85:generation,ParisBook,kittredge91:discourse}.

For those cases where the realisation remains under-determined by the
task element type, we conducted a finer-grained analysis,
by overlaying a genre partition on the  undifferentiated data.
We distinguished earlier two genres with which we are concerned:
ready-reference and step-by-step. In the manual analysed, we
recognised two more specific communicative purposes in the step-by-step
section: to enable the reader to perform a task, and to increase the
reader's knowledge about the task, the way to achieve it, or the
properties of the system as a whole.  Because of their distinct
communicative purposes, we again feel justified in calling these
genres.  We label them respectively {\em procedure\/} \ and
{\em elaboration\/}.  The intention that the reader should recognise
the differences in function of each section is underscored by the use of
distinctive typographical devices, such as fonts and
lay-out.\footnote{See \cite{drafter-wp14} for examples extracted from
the manuals.} 

The first step at this stage of the analysis was to establish whether
there was an effective overlap in task elements among the three genres
under consideration.  The results of this step is shown in
Figure~\ref{cross-ref}.  Sub-step and goal are found in all three
genres, while constraint, result and function occur in both
ready-reference and elaboration but are absent from procedure.

The next step was to undertake a  comparative analysis of  the
lexico-grammatical features  found in 
the  three genres.  This analysis indicated   that  the language employed in
these different sections of the text  varies greatly.  
We summarise here the  two genres that are
strongly contrasted:   procedure and ready-reference. Elaboration
shares features  with both  of these.
\begin{figure*}[t]
\lline 
\begin{tabbing}
nnnnnnnn\= nnnnnnnnnn\= Procedure nn \= Ready-referencennn \= Elaboration\kill 
\> \> {\bf Procedure} \> {\bf Ready-Reference} \> {\bf Elaboration} \\[0.08in]
\> Non-modal \>  100.0\% \> 75.0\% \> 72.6\% \\
\> Modal     \>    0.0\% \> 25.0\% \> 28.4\% 
\end{tabbing}
\caption{Genre-related differences in the modal system for goal}
\lline
\label{goal-cross}
\begin{tabbing}
nnnnnnnn\= nnnnnnnnn\=  Ready-referencennnnn \= Elaboration\kill 
\>  \> {\bf Ready-Reference} \> {\bf Elaboration} \\[0.08in]
\> Negative       \> 0.0\% \> 41.7\% \\
\> Positive       \> 100\% \> 58.3\%  
\end{tabbing}
\caption{Genre-related differences in the polarity system for constraint}
\lline
\label{constraint-cross}
\begin{tabbing}
nnnnnnnn\= nnnnnnnnnnnnnnnnnn\= Procedure nnn \= Ready-referencennnnnn \= Elaboration\kill 
\> \> {\bf Procedure} \> {\bf Ready-Reference} \> {\bf Elaboration} \\[0.08in]
\> Imperative     \> 97.3\%  \> 44.4\% \> 77.6\% \\
\> Declarative    \>  2.7\%  \> 55.6\% \> 22.4\% 
\end{tabbing}
\caption{Genre-related differences in the mood system for substep}
\lline
\label{substep-cross}
\end{figure*}

\begin{description}

\item [procedure:]   The top-level goal of the
user is expressed as a nominalisation.  Actions to  be achieved by the
reader are    almost exclusively  realised   by  imperatives, directly
addressing   the reader.  These  actions  are mostly material directed
actions, and there are no  causatives.  Few modals are employed,  and,
when they are, it is to express obligation impersonally.  The polarity
of processes is always positive.  Procedure employs mostly independent
clauses,  and, when clause  complexes  are used, the conjunctions  are
mostly purpose (linking  a user goal  and an action) and  alternative
(linking two user actions or two goals).

\item [ready-reference:] In this genre, all task  elements are always
realised through clauses. The declarative  mood predominates, with few
imperatives addressing the  reader.  Virtually all  the  causatives
occur here. On the dimension of modality, the  emphasis is on personal
possibility, rather than obligation, and on inclination.  We find in
this genre most of the verbal processes, entirely absent from
procedure.  Ready-reference is more weighted than procedure towards
dependent clauses, and is particularly marked by the presence of
temporal conjunctions. 
\end{description}

The  analysis  so far demonstrates  that  genre, like  task structure,
provides  some measure of  control  over the  linguistic resources but
that neither of these alone is sufficient to drive a generation system.
The final step was therefore to look at the realisations of the task
elements differentiated by genre, in cases where the realisation was
not strongly determined by the task element.  

We  refer   the   reader   back  to  Figure~\ref{task-element-realisation},  and   the
under-constrained cases of  modality  in goal, polarity  in constraint,
and mood  in substep.  Figure~\ref{goal-cross} shows  the realisations
the  task element goal with respect  to the modal system, which brings
into  sharp relief    the    absence  of modality  from     procedure.
Figure~\ref{constraint-cross} presents the   realisations by genre  of
the polarity system  for  constraint.  We  observe that only  positive
polarity occurs in ready-reference.
Finally, we note from Figure~\ref{substep-cross} that
the realisation of sub-steps is heavily loaded in favour of
imperatives in procedure.

These figures show that genre does indeed provide useful additional
control over the expression of task elements, which can be exploited by
a text generation system.  Neither task structure nor genre alone is
sufficient to provide this control, but, taken together, they offer a
real prospect of adequate control over the output of a text generator.

\section{Related Work}

The results from our  linguistic  analysis are consistent with   other
research on sublanguages  in the instructions  domain,  in both French
and English, e.g.,
\cite{kosseim-lapalme94,paris94:discourse}.  Our analysis
goes beyond previous work by identifying within the discourse
context the means for exercising explicit control over a text
generator.

An  interesting   difference with respect to previous descriptions is 
the   use of   the  true (or direct) 
imperative to express  an action in the   procedure genre, as  results
from   \cite{paris94:discourse}    seem   to  indicate     that    the
infinitive-form of the imperative  is preferred in French.  These
results, however,  were obtained from  a corpus of instructions mostly
for domestic appliances as opposed to software manuals.
Furthermore the use of the infinitive-form in instructions in general
as observed by \cite{Kocourek82} is declining, as some  of the
conventions     already common in   English  technical writing are
being adopted by French technical writers,  e.g.,
\cite{Timbal90}. 

We also note that the patterns  of realisations uncovered in our
analysis follow the principle 
of good technical writing  practice   known  as the {\em    minimalist
approach\/}, e.g.,
\cite{Carroll-94:tech-writing,Hammond94:tech-writing}.
Moreover, we observe that our corpus does not exhibit shortcomings identified 
in a Systemic Functional analysis of English software manuals
\cite{Plum90-eda-text-analysis}, such as a high incidence of agentless
passive and a failure to distinguish the function of informing from
that of instructing.

Other   work has focused  on the  cross-linguistic realisations of two
specific    semantic      relations  ({\em   generation\/}    and {\em
enablement\/})
\cite{delin94:discourse,delin96-coling}, in a  more general corpus
of  instructions 
for household appliances.  Our work focuses on the single application
domain of software instructions.  However, it takes into consideration the
whole task structure and looks at the realisation of semantic elements
as found in the knowledge base, instead of two semantic relations not
explicitly present in the underlying semantic model.

\section{Conclusion}

In this paper we have shown how genre and task structure provide two
essential sources  of 
control over the text generation process. Genre  does so by constraining
the selection of the task  elements and the range of their
expressions.  These elements, which are the procedural
representation of the user's tasks, constitute a layer of control
which mediates between genre and text, but which, without genre,
cannot control the grammar adequately.

The  work presented here  is informing  the   development of our  text
generator by  specifying the necessary  coverage of the French grammar
to  be implemented, the  required   discourse structures, and  the
mechanisms needed to control them. We continue to explore further
situational and contextual factors which might allow a system to fully
control its available linguistic resources.


\end{document}